\documentclass[letterpaper, 10 pt, conference]{ieeeconf}  

\IEEEoverridecommandlockouts                              

\usepackage{amsmath,amssymb,amsfonts}
\usepackage{graphicx} 
\usepackage[dvipsnames]{xcolor}
\usepackage{algorithm}
\usepackage{booktabs} 
\usepackage{algpseudocode}
\usepackage{bbold}
\usepackage{multirow}
\usepackage{svg}

\newcommand{\first}[1]{\textcolor{ForestGreen}{\textbf{#1}}}
\newcommand{\second}[1]{\textcolor{BurntOrange}{\textbf{#1}}}

\makeatletter
\let\NAT@parse\undefined
\makeatother
\usepackage{hyperref}
\hypersetup{
    colorlinks=true,
    linkcolor=blue,
    filecolor=magenta,      
    urlcolor=cyan,
    pdftitle={cramp},
    pdfpagemode=FullScreen,
    }

\title{\LARGE \bf
Optimizing Crowd-Aware Multi-Agent Path Finding through Local Communication with Graph Neural Networks
}

\author{Phu Pham and Aniket Bera \\
{\textit{Department of Computer Science, Purdue University, USA}}
\\\textbf{\textit{Project page:}} \href{https://ideas.cs.purdue.edu/research/projects/cramp}{https://ideas.cs.purdue.edu/research/projects/cramp}
\thanks{The authors are with the Department of Computer Science, Purdue University, USA,
        {\tt\small \{pham84,aniketbera\}@purdue.edu}}
}

\begin{document}

\maketitle
\thispagestyle{empty}
\pagestyle{empty}

\begin{abstract}

Multi-Agent Path Finding (MAPF) in crowded environments presents a challenging problem in motion planning, aiming to find collision-free paths for all agents in the system. MAPF finds a wide range of applications in various domains, including aerial swarms, autonomous warehouse robotics, and self-driving vehicles. Current approaches to MAPF generally fall into two main categories: centralized and decentralized planning. Centralized planning suffers from the curse of dimensionality when the number of agents or states increases and thus does not scale well in large and complex environments. On the other hand, decentralized planning enables agents to engage in real-time path planning within a partially observable environment, demonstrating implicit coordination. However, they suffer from slow convergence and performance degradation in dense environments. In this paper, we introduce CRAMP, a novel crowd-aware decentralized reinforcement learning approach to address this problem by enabling efficient local communication among agents via Graph Neural Networks (GNNs), facilitating situational awareness and decision-making capabilities in congested environments. We test CRAMP on simulated environments and demonstrate that our method outperforms the state-of-the-art decentralized methods for MAPF on various metrics. CRAMP improves the solution quality up to 59\% measured in makespan and collision count, and up to 35\% improvement in success rate in comparison to previous methods.

\end{abstract}

\section{INTRODUCTION}

Multi-Agent Path Finding (MAPF) poses challenges with broad applications in autonomous warehouses, robotics, aerial swarms, and self-driving vehicles \cite{MaoudjWarehouse, CLMAPF, Gameplan, ClusterSwarm, D-MUAV, HonigWarehouse, patelDronerf}. The objective is to plan paths for multiple agents to navigate from start to goal positions in obstacle-laden environments. A critical constraint of such systems is to ensure agents navigate concurrently without collisions. Two main categories of approaches are centralized and decentralized planning. Centralized planners \cite{FransenAstar, ODrMstar, CBS, Okumura2022lacam} aim to find optimal solutions that minimize the length of collision-free paths. They are effective in small and sparse environments but face limitations in real-time performance and scalability in large and dense environments \cite{LamBCP, LiLNS}. These methods require complete knowledge of the environment and full replanning when a change occurs, leading to exponential computation times with increased agents, obstacles, and world size. Recent studies \cite{OkumuraRefinement, Okumura2022lacam} have sought to discover real-time solutions. However, these solutions remain sub-optimal and still require access to global information about the world.

On the contrary, decentralized methods \cite{PRIMAL, PRIMAL2, PICO, CPL, ChenContrastive, MaoudjWarehouse} seek to tackle these challenges by allowing each agent to acquire their own policies. In these approaches, agents can reactively plan paths within partially observable environments. Such methods prove beneficial in situations where agents lack comprehensive knowledge of the world, as is often the case in the domain of autonomous vehicles. Rather than pursuing an optimal solution for all agents, decentralized planners train local policies that rapidly generate sub-optimal solutions as a tradeoff between speed and solution quality. Given that agents make their decisions based on local information, decentralized approaches often face challenges in achieving effective global coordination among agents. In cluttered, dynamic environments characterized by congestion or rapid changes, agents may tend to prioritize their individual goals, potentially resulting in conflicts and inefficiencies that affect the overall system's performance.

\begin{figure}

    \centering
    \includegraphics[width=\columnwidth]{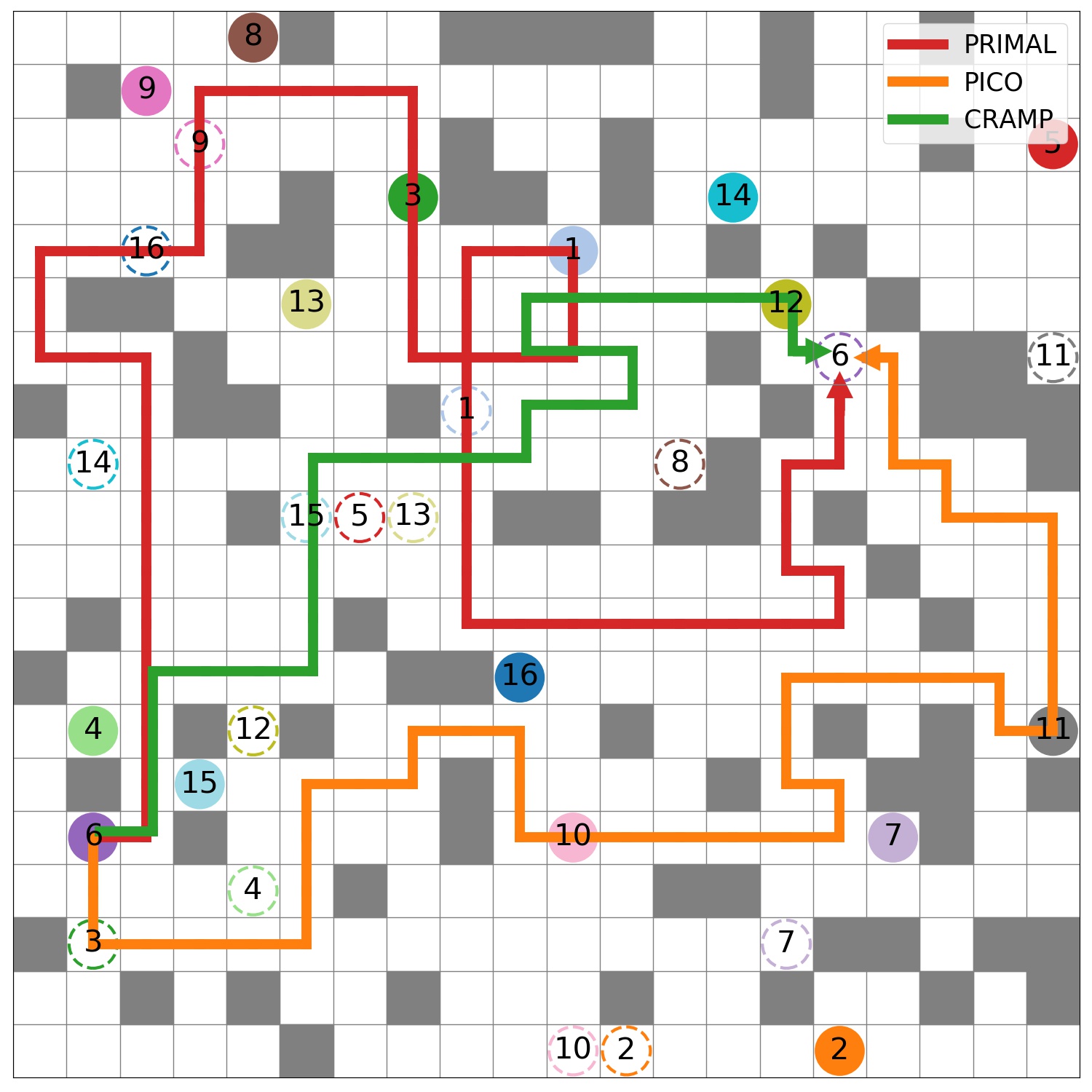}
    \caption{Comparison of paths between our CRAMP approach and the previous state-of-the-art methods. Colored circles with numerical labels represent the starting positions of agents, while dashed circles with corresponding numbers denote the target positions. The obstacles are marked with grey squares. Our innovative crowd-aware method for multi-agent path-finding challenges significantly outperforms existing approaches such as PRIMAL \cite{PRIMAL} and PICO \cite{PICO}, yielding notably shorter solutions.}
    \label{fig:overview}
    \vspace{-20pt}
\end{figure}


To tackle the challenges presented by dense crowds, our approach, CRAMP, integrates a crowd-awareness mechanism and local communication via Graph Neural Networks (GNN) \cite{GNNKipf}. This strategy is guided by a boosted curriculum training approach, which gradually exposes agents to increasingly complex scenarios. CRAMP is specifically designed to train intelligent agents capable of efficiently navigating in dense and dynamic environments. By formulating the MAPF problem as a sequential decision-making task, we employ deep reinforcement learning techniques, enabling agents to make optimal decisions while considering the presence of other agents and the dynamic nature of the environment.

We evaluate the CRAMP approach through extensive experiments on various synthetic environments, comparing it against state-of-the-art MAPF algorithms. Our results demonstrate that CRAMP achieves superior performance in terms of solution quality and computational efficiency while maintaining robustness in crowded settings. Moreover, we showcase the adaptability of our approach in dynamic and evolving environments. Figure \ref{fig:overview} illustrates a comparison of computed paths between PRIMAL \cite{PRIMAL}, PICO \cite{PICO}, and our method, CRAMP. Our approach significantly shortens the path to the target position while mitigating potential congestion by prioritizing actions that lead to less densely populated areas. This strategy effectively reduces the likelihood of deadlocks and enhances overall path efficiency.

Our main contributions can be summarized as follows:
\begin{itemize}
    \item We present a reinforcement learning approach for the multi-agent path-finding problem, utilizing a carefully designed crowd-aware mechanism that enables agents to adapt their policies dynamically based on crowd density and behavior.
    \item Exploiting local communication between nearby agents via Graph Neural Networks, we facilitate agents to gather information from and predict the behavior of nearby agents, thereby enhancing their decision-making capabilities.
    \item We conduct a series of experiments, conclusively showcasing that our approach surpasses state-of-the-art methods across a range of metrics, including success rate, collision count, and makespan.
\end{itemize}

\section{RELATED WORK}

Researchers have developed a wide range of algorithms and techniques to solve the MAPF problem, aiming to strike a balance between optimality, speed, and adaptability to varied maps and conditions. Centralized approaches have traditionally been favored for MAPF, but they come with several drawbacks. As the number of agents and the complexity of the environment increase, centralized algorithms may become computationally intractable, leading to long planning times \cite{CBS, foster2023complexity}. Furthermore, they are susceptible to single points of failure, where if the central planner fails, the entire system may collapse. 

Decentralized planners have garnered attention due to their ability to mitigate the computational complexity associated with centralized planners and their resilience against single points of failure. However, these approaches may face challenges in managing communication overhead and ensuring effective coordination, especially in scenarios with a large number of agents \cite{WSCaS, DeMDP, PRIMAL}.

Moreover, the exploration of hybrid techniques that integrate both centralized and decentralized components remains an important avenue of research \cite{AlexeyHybrid, DRLReview}. Hybrid approaches could potentially capitalize on the strengths of both paradigms while mitigating their respective limitations. In this work, we primarily study the centralized and decentralized approaches. Future research could explore hybrid approaches to gain a more comprehensive understanding of the tradeoffs involved and to further advance the field.

\subsection{Centralized approaches}
Centralized approaches for MAPF involve using a centralized planner to find optimal or near-optimal paths for all agents simultaneously. These approaches are suitable for scenarios where global coordination among agents is feasible and desired. Among the centralized methods, Conflict-based search (CBS) \cite{CBS} is a widely adopted algorithm that decomposes the MAPF problem into single-agent path-finding problems while considering agent conflicts. It iteratively identifies conflicts, prioritizes them, and resolves them by searching for alternative paths or time steps for the conflicting agents. Other variants of CBS \cite{iCBS, imprCBS, hCBS, EECBS} extend CBS to handle dynamic environments and continuous time. 

Extending A* \cite{Astar} to MAPF, hierarchical A* \cite{hAstar} and M* \cite{Mstar} plan individual paths for each agent and then search forward in time to detect potential collisions. When collisions occur, they perform joint planning through limited backtracking to resolve the collision and continue with individual agent plans. One variant, ODrM* \cite{ODrM*}, reduces the need for joint planning by breaking down the problem into independent collision sets and employs Operator Decomposition (OD) \cite{OD} to manage search complexity effectively.

Another line of approach is sampling-based methods \cite{RRT, PRM, RRTstar}, which have also been adapted to MAPF. These methods generate random samples of possible agent configurations and construct paths by connecting these samples while avoiding collisions. 

Recently, MAPF-LNS \cite{LiLNS} and MAPF-LNS2 \cite{LNS2} solvers can achieve near-optimal solutions in a short amount of time using Large Neighborhood Search. LaCAM \cite{Okumura2022lacam} proposes a two-level search algorithm to solve large MAPF instances with up to thousands of agents in a matter of seconds. All the above-mentioned approaches require global information.

\subsection{Decentralized approaches}

Recent studies have directed their attention towards decentralized policy learning \cite{DeMDP, PRIMAL, WSCaS, PICO, CPL, ChenContrastive, DistHeuMAPF}, in which individual agents learn their own policies. \cite{DeMDP} models the MAPF problem as a Partially Observable Markov Decision Process (POMDP) and uses hierarchical graph-based macro-actions to produce robust solutions. A macro-action is a higher-level action composed of a sequence of atomic actions that an agent can take in its environment.

Additionally, \cite{WSCaS} proposes a method called WSCaS (Walk, Stop, Count, and Swap) that optimizes the initial solutions provided by A*. Each agent $a_i$ initially plans a path $\mathcal{P}_i = \{p_0, ..., p_n\}$ from the start position $s_i$ to the goal position $g_i$ using the A* algorithm. At each position $p_j$ on the path $\mathbf{P}_i$, the agent $a_i$ utilizes local communication to reactively adjust the path according to the current environment to avoid collisions and deadlocks until it reaches $g_i$.

In contrast to conventional methods, \cite{MAPFGame} formulates the MAPF problem within a game theory setting, where each agent tries to maximize their utility function while respecting safety constraints. In this framework, each agent simulates a good learned policy while avoiding the bad ones.

\subsection{Multi-agent reinforcement learning}

The trajectory of multi-agent reinforcement learning (MARL) research features key contributions that have systematically addressed the complexities of agent coordination and navigation. PRIMAL \cite{PRIMAL} focuses on integrating reinforcement learning with imitation learning to enhance local policy development and inter-agent coordination. PICO \cite{PICO} introduced prioritized communication learning methods, optimizing information exchange and decision-making processes among agents. Additionally, extensions like CPL \cite{CPL} have adopted a curriculum learning framework, applying distinct loss functions at varying curriculum levels to progressively refine agent policies and tackle the challenges of learning in highly dynamic settings.

\subsection{Graph neural networks for MAPF}

Graph Neural Networks (GNNs) have gained significant attention and adoption within multiagent systems due to their demonstrated efficacy in various tasks \cite{LiGnn, RAIST, nayak22informarl, patel2023dream, peng2023graphbased}. In the MAPF settings, GNNs offer a compelling solution by facilitating complex interaction modeling, local communication, and adaptation to dynamic environments. By representing agents and their surroundings as nodes and edges in a graph structure, GNNs enable efficient coordination and decision-making among agents.

Despite the promising potential, limited work exists in applying GNNs to MAPF. Notably, \cite{LiGnn} integrated GNNs into MAPF but trained the model through imitation learning, leading to an imbalance between exploration and exploitation. As a consequence, the model's generalization in new environments is compromised. In this work, we aim to address this issue by integrating GNNs into our framework, seeking to enhance the generalization capabilities of our model. Our contribution to this landscape includes a novel crowd-avoidance mechanism supported by local communication through GNNs. We adopt a simple yet effective curriculum-driven training strategy to enhance learning efficiency and significantly reduce incidents of collisions and deadlocks. 


\begin{figure}
\centerline{\includegraphics[width=.99\columnwidth]{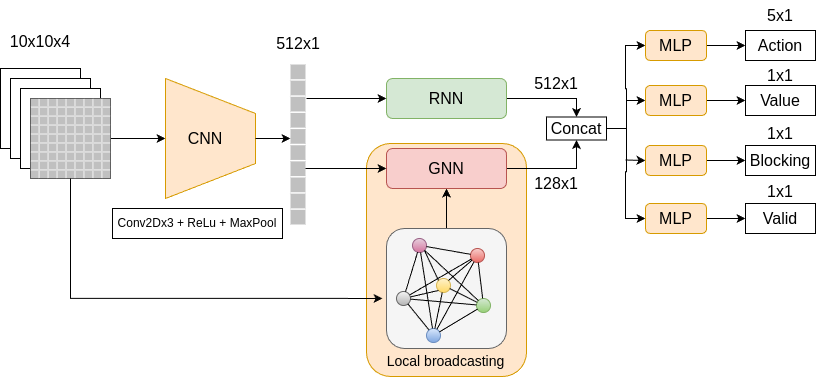}}
\caption{The network architecture to train the distributed local policies. The network takes local observation of size $10 \times 10 \times 4$ as inputs. The network's outputs include the predicted probabilities for the actions (size $5 \times 1$), predicted return value, validity of predicted action and blocking probabilities (all of size $1\times1$).}
\label{fig:acnet}
\vspace{-15pt}
\end{figure}

\section{OUR APPROACH}\label{sec:method}

This section outlines CRAMP, our approach to addressing the MAPF problem in densely populated environments. The methodology can be divided into several key components, including world modeling, policy learning, a crowd-aware reward function, GNN-based local communication, and a boosted curriculum training strategy.

\subsection{World Modeling}
Like most reinforcement learning techniques, we formulate the MAPF problem in a discrete grid-world environment. In this setting, agents partially observe the world within a limited local region of size $m \times m$ centered around themselves. Within this local region, each agent can perceive its own goal, the presence of nearby agents, and the directions to those agents' goals. These observations are used to train the local policy network.

The action space of each agent includes five discrete actions in the grid world: moving one cell in one of the four directions (N, E, S, W) or remaining stationary. An action is considered valid if it avoids collisions with obstacles and other agents or goes beyond the world's boundaries.

\subsection{Reinforcement learning for local policy training}

We train the agent's policy using deep reinforcement learning and demonstration learning techniques. Multiple agents are independently trained in a shared environment utilizing the Asynchronous Advantage Actor-Critic (A3C) algorithm \cite{A3C}. The inputs to the policy network comprise four layers of an agent's local observation, which include the grid world, obstacles, neighboring agents, and neighbors' goals, alongside its own goal position. The network's outputs consist of the next action, predicted return value, validity of the action, and blocking probability. The detailed architecture of the policy network is illustrated in Fig. \ref{fig:acnet}. The CNN feature extractor employs two sequential blocks of [Conv2D-Conv2D-Conv2D-Relu-MaxPool]. The CNN's output is a $512 \times 1$ vector. This feature vector is subsequently input into an LSTM to encode temporal features. In the GNN block, the adjacency matrix is computed based on the local field of view (FOV), where the ego agent is connected to all other agents within the FOV. For each agent within the ego agent's FOV, encoded feature vectors are extracted to create a node embedding matrix of size $N \times 512$, where $N$ represents the maximum number of visible agents. The GNN processes both the adjacency and node embedding matrices to produce an $N \times 128$ matrix, with each row reflecting a graph signal from each agent. These nodes are averaged and concatenated with the LSTM feature vector. The resulting concatenated feature vector is fed into fully-connected layers to predict the variables.

These outputs are updated after every batch of $T=256$ steps or after each episode. An episode is finished when all agents successfully arrive at their goals or when the maximum number of episode steps has been reached.

Consider a timestep \( t \); the network aims to maximize the discounted return for the trajectory starting at \( t \):
\vspace*{-5pt}
\begin{equation}
    \mathcal{R}_t = \sum_{i=0}^\infty \gamma^i r_{t+i}
\end{equation}
where \( \gamma \) is the discount factor and \( r_{t+i} \) is the reward received by the agent at timestep \( t+i \) after taking a series of actions. The optimization can be achieved by minimizing the loss function:
\vspace*{-5pt}
\begin{equation}
    \mathcal{L}_{\text{value}} = \sum_{t=0}^T(V(o_t) - \mathcal{R}_t)^2
\end{equation}
where \( V(o_t) \) is the predicted value of the policy network at timestep \( t \).

The policy is updated by adding the entropy term \( H = -\sum_{a} \pi(a_t = a | o_t) \log(\pi(a_t = a | o_t)) \) \cite{SAC} to the policy loss:
\vspace*{-5pt}
\begin{equation}
    \mathcal{L}_\pi = \epsilon H - \sum_{t=0}^T \log(P(a_t | \pi, o) A (o_t, a_t))
\end{equation}
in which \( \epsilon \) is the small weight for the entropy term \( H \), \( P(a_t | \pi, o) \) is the probability of taking action \( a \) at timestep \( t \) with observation \( o \) by following the policy \( \pi \), and \( A (o_t, a_t) \) is the approximation of the advantage function.
\begin{equation}
    A(o_t, a_t) = \sum_{i=0}^{k-1}\gamma^i r_{t+i} + \gamma^k V(o_{k+t}) - V(o_t)
\end{equation}

An agent \( a_i \) is considered blocking another agent \( a_j \) if \( a_i \) is on the optimal path of \( a_j \) to the goal from its current position and forces \( a_j \) to take a detour at least ten steps longer. The optimal paths of these agents are determined using the \( A^* \) algorithm. The \( \textbf{\textit{blocking}} \) output is used to predict the blocking probability. We use binary cross-entropy loss as the loss function for this term.
\begin{equation}
    \mathcal{L}_{\text{block}} = - \sum_{i=1}^{T} y_i \log(p_{i}) + (1 - y_i) \log(1 - p_i)    
\end{equation}

where \( y_i \) represents the ground truth binary label indicating whether the ego agent is blocking any other agents, \( p_i \) is the predicted blocking probability. 

The same binary cross-entropy loss is applied to the action validity loss \( \mathcal{L}_{\text{valid}} \). The total loss is then formulated as the sum of all losses:
\begin{equation}
    \mathcal{L}_{\text{total}} = \mathcal{L}_{\pi} + \mathcal{L}_{\text{value}} + \mathcal{L}_{\text{blocking}} + \mathcal{L}_{\text{valid}}
\end{equation}

\subsection{Demonstration learning} \label{sec:demonstration}

Previous studies \cite{DemonstrationStefan, ImitationNair, Skild2021, PRIMAL} show that learning from demonstrations can facilitate agents to confine the high-reward regions quickly. The agents then use reinforcement learning to explore these regions to improve their policies. Adopting this approach from PRIMAL \cite{PRIMAL}, we also use ODrM* \cite{ODrM*} as the expert agent. The agents will first learn from the ODrM* expert and then randomly switch between exploration and demonstration modes to update their local policies.
 
\subsection{Crowd-aware reward function}

In reinforcement learning for grid-world MAPF, the reward system is simple yet effective: agents lose points for each timestep away from their goal, motivating them to find the fastest route. They are also penalized for idleness before reaching their goal, encouraging exploration and action. Major penalties are given for collisions, exiting the environment, or hitting obstacles, discouraging such actions.

Decentralized policies might result in agents behaving selfishly, blocking others from their goals, and potentially causing deadlocks in tight areas. To foster teamwork, agents receive a significant reward when all complete an episode together but face penalties for hindering others, promoting cooperation and efficiency in the multi-agent system.

Even though these rewards are widely used in MAPF, they often overlook the issue of traffic congestion. None of the previously mentioned reward terms consider agents' behavior in densely populated areas. Typically, deadlocks often happen in congested regions, where the available moving space is severely limited. To reduce the likelihood of a deadlock, we introduce an innovative reward term, denoted as $R_n$, that could boost agents' performance in terms of deadlock rates. We define an event $\mathcal{E} = \{\delta_{a} \geq \zeta \}$, where $\delta_{a}$ is the density of agents close to agent $a$ and $\zeta$ is a threshold. The density $\delta_{a}$ is calculated by counting the number of agents in $a$'s field of view of size $w$ divided by the free space in that region. Notice that this region is more confined than the partially observable world ($w = 5$ in our setup). If event $\mathcal{E}$ occurs, we consider the area as $\zeta$-crowded; thus, the probability of deadlock significantly increases. Consequently, when an agent transitions from a non-crowded region to a $\zeta$-crowded one, it incurs a negative reward. Conversely, when an agent moves out of a $\zeta$-crowded region, it receives a positive reward. Moving between the non-crowded areas or $\zeta$-crowded areas has no impact on the reward.

The threshold $\zeta$ is not fixed; its magnitude should be higher when the environment contains more agents and obstacles and lower in sparser environments. We define $\zeta$ as follows:
\vspace{-5pt}
\begin{equation}
    \zeta = min\bigg(1.0, 0.7 + \frac{A}{m^2(1 - d) - A}\bigg)
\end{equation}
\vspace{-1pt}
where $A, m, d$ are the number of agents, grid world size, and obstacle density, respectively. The value $0.7$ represents the base level of congestion tolerance. The addition of $A / (m^2(1 - d) - A)$ dynamically adjusts this tolerance based on the number of agents and available space of the environment. In dense environments, the threshold becomes more lenient (i.e., higher $\zeta$) but is capped at $1.0$ to prevent it from becoming too lenient, thus maintaining a level of challenge in navigation and preventing overly dense agent clustering. We conducted tests on the lower and upper bounds within the range of $0.5$ to $1.5$ and determined that the range between $0.7$ and $1.0$ yields the best results.

\begin{figure}
    \centering
    \includegraphics[width=.99\columnwidth]{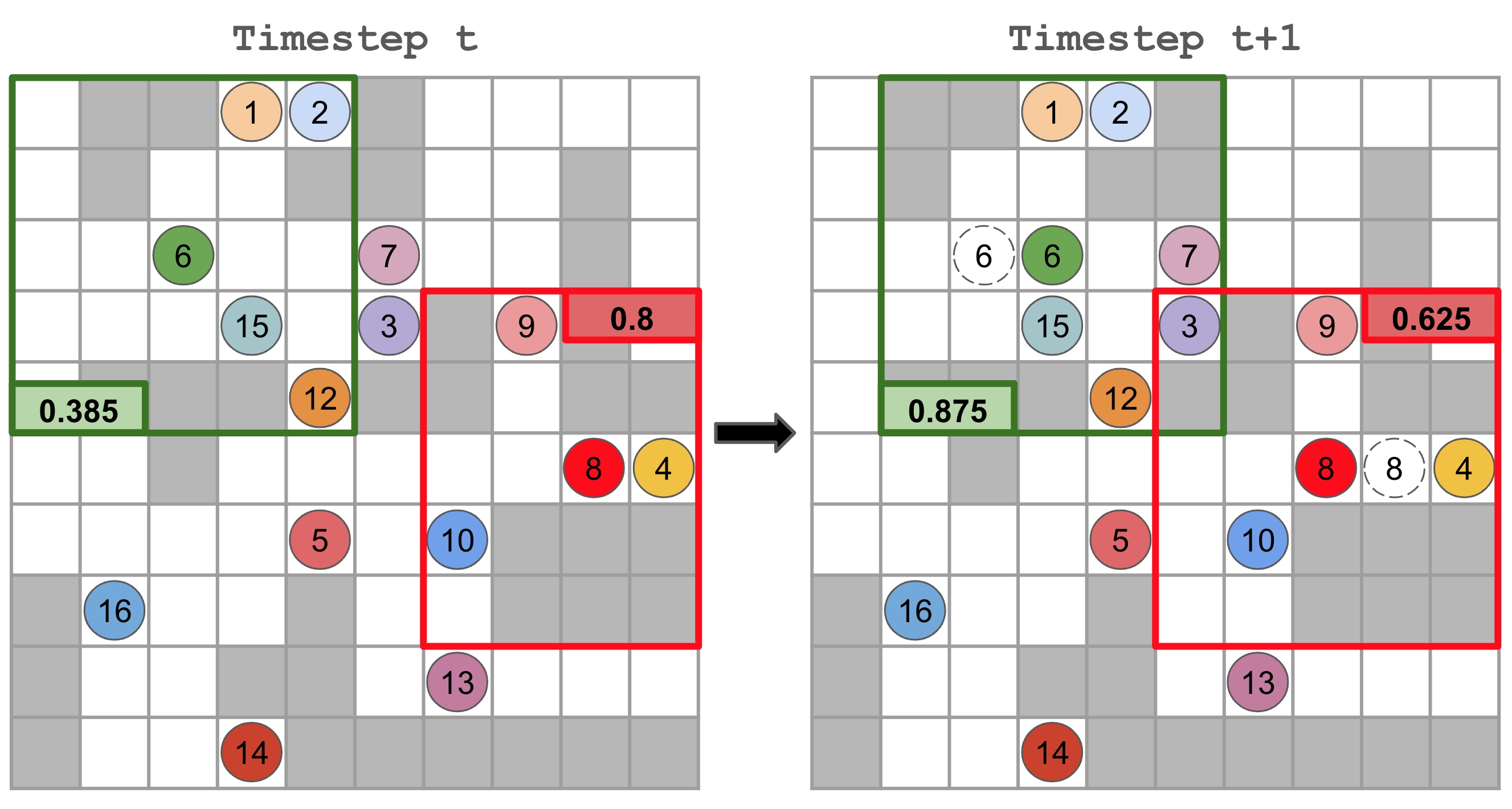}
    \caption{Example of agents getting negative and positive rewards by moving in or out of a $\zeta$-crowded region. The colored circles with numerical labels are the agents and the dashed circles represent the previous positions.}
    \label{fig:crowdreward}
    \vspace*{-15pt}
\end{figure}

Fig. \ref{fig:crowdreward} showcases two scenarios where the agents get a positive reward by moving out and a negative reward by moving in a $\zeta$-crowded region. In the first scenario, consider agent number $6$ (green agent) at timestep $t$ with the field of view inside the green border; the agent is at the sparse region where the crowd density is $d = 5 / 13 = 0.385$. If the agent moves to the right, where the crowd density is much higher $d = 7 / 8 = 0.875$, the agent will be punished with a reward of $-0.5$. On the contrary, in the second scenario with the agent number $8$ (red agent), if the agent moves to the left, the crowd density reduces from $d = 4 / 5 = 0.8$ to $d = 5 / 8 = 0.625$, and the agent receives a positive reward of $+0.5$. We assume the threshold $\zeta = 0.75$ in both cases.

Our reward function can be formulated as:
\begin{equation} \label{rewardVector}
R=
\begin{bmatrix}
     R_m & R_c & R_s & R_e & R_n
\end{bmatrix}
\end{equation}
\vspace{-8pt}
\begin{equation} \label{weightVector}
w=
\begin{bmatrix}
     w_m & w_c & w_s & w_e & w_n
\end{bmatrix}^T
\end{equation}
\vspace*{-17pt}
\begin{equation} \label{totalReward}
    R_{total} = R \cdot w
\end{equation}

In \eqref{rewardVector}, $R_m = -0.3$ is the reward for the agent's movement in the four direction (N/E/S/W). $R_c = -2.0$ is the reward for agent collisions. $R_s$ is the reward if the agent is standing still. $R_s = 0$ if the agent is at its goal, $R_s = -0.5$ otherwise. $R_e = +20.0$ is the team reward given to all agents when all agents reach their goals. $R_n$ is the crowd-aware reward. $R_n = +0.5$ if the agent moves out of a $\zeta$-crowded region, $R_n = -0.5$ if the agent moves in a $\zeta$-crowded one. $R_n = 0$ otherwise.
In \eqref{weightVector}, $w_m, w_c, w_e, w_n$ are the counts of each reward type in an episode.

\subsection{Graph neural network for local communication}

We utilize Graph Neural Networks (GNNs) to facilitate local communication within a multiagent system, leveraging the intrinsic structure and relationships among nearby agents. The GNN architecture is specifically tailored to model the spatial relationships among agents by treating each agent as a node within a graph structure, $G = (V, E)$, where $V$ denotes nodes (agents) and $E$ represents edges (proximities between agents). The graph's adjacency matrix, $A$, is dynamically computed from the local view, with edge weights, $A_{ij}$, encoding the distances between agents. These weights control the flow of information between nodes, determining how much influence each node has on its neighbors during updates. Each node $i$ is associated with a feature vector $x^i$, derived from the CNN's output, representing the agent's observed state.

The GNN architecture consists of \( L \) layers, with each layer \( l \) outputs a graph signal represented by \( F_l \) features. This input signal is fed to the next layer and then linearly transformed by the graph filters \( H_l \) to leverage the graph's structure, represented by the adjacency matrix \( A \). The output features \( F_{l+1} \) are then passed through an activation function \( \sigma_{l+1} \). Specifically, for a given layer \( l \), the output is given by:

\[ h_{l+1}^f = \sigma_{l+1} \left( \sum_{g=1}^{F_{l}} H_{l}(A) h_l^g \right) \]

Here, \(h_l^i\) is node \( i \)'s input signal (\(h_0^i = x^i\)), \( H_l \) are the linear operators at layer \(l\) that leverage the graph structure for processing, usually by relying solely on local neighbor exchanges and partial information about the graph.

The final layer's output, $h_L^i$, encapsulates aggregated neighborhood information, emphasizing the importance of spatial relationships for informed decision-making in multi-agent systems. 

\subsection{Boosted curriculum learning} \label{sec:curriculum}

We adopt a curriculum training strategy by gradually exposing the agents to increasingly complex settings over time. Initially, the agents are trained in small and sparse environments. If the agent's performance, measured by total rewards earned, does not show improvement after $n$ consecutive episodes or after a set number of training episodes ($50k$ in our case), we escalate the complexity of the environment. Since the environment size and density are randomly sampled from a range after each training episode, we expand these ranges at each curriculum level.

\begin{align*}
    d_l, d_h &= (\min(0.05 \sigma, 0.2), \min(0.1 + 0.1 \sigma, 0.6)) \\
    s_l, s_h &= (\min(10 + 5\sigma, 40), \min(40 + 5 \sigma, 120)) \\
    n &= n \times 1.5 
\end{align*}
Here, $\sigma$ denotes the current curriculum level, and $d_l$, $d_h$, $s_l$, and $s_h$ represent the ranges of obstacle density and world size to sample from.

Additionally, our approach incorporates a boosting technique. When agents exhibit poor performance in specific environments, we address this by re-sampling those environments and subsequently retraining the agents. This strategy aims to enhance the agents' proficiency in challenging scenarios by providing them with additional exposure and training opportunities in such environments.

\begin{table*}
\centering
\caption{Performance results of the CRAMP model assessed in varied environments with dimensions $20\times20$, agent counts ranging from 8 to 64, and obstacle densities varying from 0 to 0.3, specified right below the metrics. \first{Green} and \second{orange} colors indicate the best and second best results. The hyphen ( - ) symbol denotes invalid entries due to a 0 success rate.}
\label{table:results}
\resizebox{\textwidth}{!}{%
\begin{tabular}{|c|cccc|cccc|cccc|cccc|}
\hline
\multirow{3}{*}{Methods} & \multicolumn{16}{c|}{\textbf{8 agents}} \\
\cline{2-17}
& \multicolumn{4}{|c|}{\textbf{Success rate}} & \multicolumn{4}{|c|}{\textbf{Makespan}} & \multicolumn{4}{|c|}{\textbf{Total moves}} & \multicolumn{4}{|c|}{\textbf{Collision count}}\\
& 0  & 0.1 & 0.2 & 0.3 & 0 & 0.1 & 0.2 & 0.3 & 0 & 0.1 & 0.2 & 0.3 & 0 & 0.1 & 0.2 & 0.3\\
\hline
PRIMAL & \second{93}  & 90  & 48 & 15 & 35  & 63  & 149 & 234 & 221  & 223  & 345  & 565  & 1.94   & 3.02   & 3.03  & 5.98  \\
DHC    & 91  & 87  & 55 & 11 & 34  & 63  & 137 & 242 & 272  & 240  & 401  & 639  & 1.66   & 2.72   & 3.74  & 4.17\\
PICO   & \first{100} & 96  & 55 & 25 & \second{27}  & 42  & 135 & 205 & 124  & 143  & 290  & 463  & 0.59   & \second{0.62}   & 1.31  & \second{2.32}  \\
CPL    & \first{100} & \second{99}  & \second{94} & \first{65} & \first{25}  & \second{31}  & \second{45}  & \second{124} & \first{111}  & \second{121}  & \second{150}  & \second{282}  & \first{0.30}   & \first{0.60}   & \second{1.20}  & 3.20 \\
CRAMP  & \first{100} & \first{100} & \first{95} & \second{64} & \second{27}  & \first{27}  & \first{40}  & \first{55}  & \second{117}  & \first{117}  & \first{144}  & \first{156}  & \second{0.48}   & \first{0.60}   & \first{0.90}  & \first{1.27} \\
\hline
\multirow{3}{*}{Methods} & \multicolumn{16}{c|}{\textbf{16 agents}} \\
\cline{2-17}
& \multicolumn{4}{|c|}{\textbf{Success rate}} & \multicolumn{4}{|c|}{\textbf{Makespan}} & \multicolumn{4}{|c|}{\textbf{Total moves}} & \multicolumn{4}{|c|}{\textbf{Collision count}} \\
& 0 & 0.1 & 0.2 & 0.3 & 0 & 0.1 & 0.2 & 0.3 & 0 & 0.1 & 0.2 & 0.3 & 0 & 0.1 & 0.2 & 0.3 \\
\hline
PRIMAL & 92  & 88  & 50 & 3  & 57  & 72  & 176 & 249 & 482  & 510  & 766  & 1396 & 6.59   & 8.29   & 11.63 & 17.64 \\
DHC    & \second{94}  & 88  & 48 & 5  & 54  & 71  & 169 & 242 & 477  & 477  & 822  & 1504 & 7.28   & 8.95   & 11.75 & 18.70 \\
PICO   & \first{100} & \second{95}  & 57 & 7  & 31  & 49  & 145 & 240 & 251  & 299  & 526  & 1292 & 2.98   & 3.98   & \second{4.96}  & \second{8.00}\\
CPL    & \first{100} & \second{95}  & \second{81} & \second{22} & \first{27}  & \second{41}  & \second{84}  & \second{213} & \first{221}  & \first{249}  & \second{374}  & \second{780}  & \second{2.80}   & \second{3.70}   & 5.30  & 16.10 \\
CRAMP  & \first{100} & \first{98}  & \first{83} & \first{23} & \second{30}  & \first{33}  & \first{45}  & \first{88}  & \second{236}  & \second{253}  & \first{275}  & \first{388}  & \first{2.60}   & \first{3.20}   & \first{3.70}  & \first{5.73} \\
\hline
\multirow{3}{*}{Methods} & \multicolumn{16}{c|}{\textbf{32 agents}} \\
\cline{2-17}
& \multicolumn{4}{|c|}{\textbf{Success rate}} & \multicolumn{4}{|c|}{\textbf{Makespan}} & \multicolumn{4}{|c|}{\textbf{Total moves}} & \multicolumn{4}{|c|}{\textbf{Collision count}}\\
& 0 & 0.1 & 0.2 & 0.3 & 0 & 0.1 & 0.2 & 0.3 & 0 & 0.1 & 0.2 & 0.3 & 0 & 0.1 & 0.2 & 0.3 \\
\hline
PRIMAL & \second{92}  & 72  & 9  & 0  & 54  & 108 & 245 & -   & 958  & 1094 & 2227 & -    & 26.23  & 30.48  & 47.27 & -  \\
DHC    & \second{92}  & 62  & 3  & 0  & 49  & 136 & 243 & -   & 957  & 1145 & 2009 & -    & 27.04  & 34.79  & 49.29 & -  \\
PICO   & \first{100} & 75  & 19 & 0  & 38  & 97  & 225 & -   & 551  & 774  & 1713 & -    & 14.80  & 20.62  & 36.28 & -  \\
CPL    & \first{100} & \first{92}  & \first{50} & 0  & \first{32}  & \second{58}  & \second{159} & -   & \first{471}  & \second{564}  & \second{1032} & -    & \second{11.90}  & \second{17.40}  & \second{30.30} & -  \\
CRAMP  & \first{100} & \second{82}  & \second{34} & \first{4}  & \second{37}  & \first{47}  & \first{85}  & \first{170} & \second{489}  & \first{552}  & \first{754}  & \first{1130} & \first{11.09}  & \first{16.52}  & \first{23.37} & \first{26.50} \\
\hline
\multirow{3}{*}{Methods} & \multicolumn{16}{c|}{\textbf{64 agents}} \\
\cline{2-17}
& \multicolumn{4}{|c|}{\textbf{Success rate}} & \multicolumn{4}{|c|}{\textbf{Makespan}} & \multicolumn{4}{|c|}{\textbf{Total moves}} & \multicolumn{4}{|c|}{\textbf{Collision count}} \\
& 0 & 0.1 & 0.2 & 0.3 & 0 & 0.1 & 0.2 & 0.3 & 0 & 0.1 & 0.2 & 0.3 & 0 & 0.1 & 0.2 & 0.3 \\
\hline
PRIMAL & 75  & 7   & 0  & 0  & 111 & 242 & -   & -   & 2419 & 3680 & -    & -    & 115.79 & 171.31 & -     & -   \\
DHC    & 72  & 0   & 0  & 0  & 109 & -   & -   & -   & 2121 & -    & -    & -    & 106.78 & -      & -     & -     \\
PICO   & \first{83}  & 13  & 0  & 0  & 94  & 225 & -   & -   & 1473 & 2621 & -    & -    & 90.95  & 128.40 & -     & -   \\
CPL    & \second{80}  & \second{20}  & 0  & 0  & \second{92}  & \second{128} & -   & -   & \first{1230} & \second{2204} & -    & -    & \second{84.00}  & \second{109.00} & -     & -     \\
CRAMP  & 78  & \first{27}  & \first{2}  & 0  & \first{90}  & \first{125} & \first{278} & -   & \second{1383} & \first{1877} & \first{2762} & -    & \first{80.92}  & \first{104.25} & \first{159.00} & -   \\
\hline
\end{tabular}
}
\end{table*}

\section{Experiments and results}
In this section, we present our model's outcomes in comparison to the leading multi-agent reinforcement planners, evaluated across various grid world environments.

\subsection{Experiment setup}
Our experiments were conducted on randomly generated environments featuring varying obstacle densities (0, 0.1, 0.2, 0.3) and numbers of agents (8, 16, 32, 64).

Our evaluation involves testing our model on 100 distinct environments for each specific configuration, including varying numbers of agents and obstacle densities. The final results are computed as an average across these 100 environments.

\subsection{Evaluation metrics}

We evaluate the performance of our model using the following metrics:
\begin{itemize}
    \item \textbf{Success rate}: This metric quantifies the ratio between successfully solved scenarios and the total number of tested environments. A higher success rate indicates a greater proficiency in finding solutions.
    \item \textbf{Makespan}: The makespan signifies the time steps required for all agents to reach their respective goals. It is equivalent to the longest path among all agents. Smaller makespan values indicate more efficient paths and, therefore, higher performance.
    \item \textbf{Total moves}: This metric captures the total number of non-idle actions all agents take. It provides a measure of the overall performance and coordination of the agent team.
    \item \textbf{Collision count}: The collision count represents the number of collisions encountered, including collisions with obstacles, other agents, or instances of agents moving beyond the environment boundaries.
\end{itemize}

Regarding all of these metrics, except for the success rate, smaller values indicate better performance.

\subsection{Result analysis}

The results presented in Table \ref{table:results} illustrate a comprehensive evaluation of our model's capabilities in navigating complex multi-agent environments, highlighting its strengths in terms of success rates, path quality, team performance, and collision avoidance. To ensure a fair comparison with other benchmarking methods such as PRIMAL \cite{PRIMAL}, DHC \cite{DistHeuMAPF}, PICO \cite{PICO}, and CPL \cite{CPL}, we maintain a fixed world size of $20 \times 20$ at test time. It is important to note that we were unable to reproduce the results reported in these papers due to implementation errors confirmed by the authors or because the code was not published. This limitation has influenced our ability to perform a direct comparison, and as such, the results for those methods are taken directly from PICO \cite{PICO} and CPL \cite{CPL} papers.

As depicted in Table \ref{table:results}, our model consistently demonstrates better performance across most environments, particularly in terms of success rate. Notably, our approach can find solutions in two densely populated scenarios, featuring 32 agents with a 0.3 density and 64 agents with a 0.2 density, where all other methods encounter complete failure. Across the remaining configurations, our model consistently either secures the top position or closely follows the best-performing method.

Our model ranks among the top performers when assessing solution quality metrics such as makespan and total moves. In highly congested environments, such as those with 16 agents with a 0.3 density and 32 agents with a 0.2 density, our solutions prove to be more efficient by cutting the makespan and total moves in half.

Regarding collision avoidance, our solutions consistently yield the lowest collision counts, except for the simplest environment, where all models perform equally well. It is important to note that our collision count considers all types of collisions, in contrast to other methods that only consider collisions between agents. 

\begin{figure}[ht!]
    \centering
    \begin{minipage}{0.49\columnwidth}
        \includegraphics[width=\linewidth]{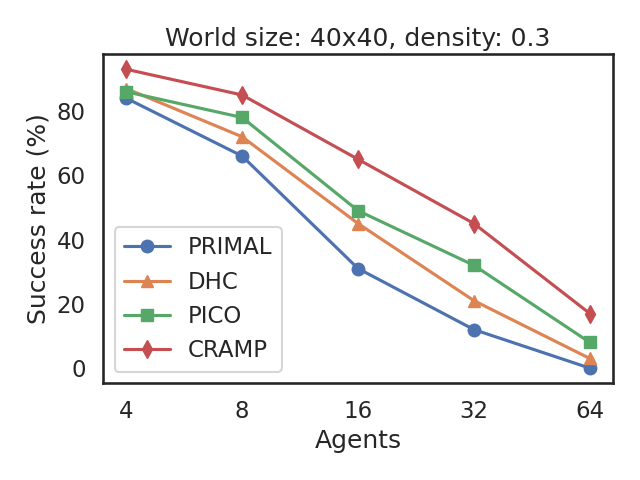}
        \label{fig:world40}
    \end{minipage} 
    \begin{minipage}{0.49\columnwidth}
        \includegraphics[width=\linewidth]{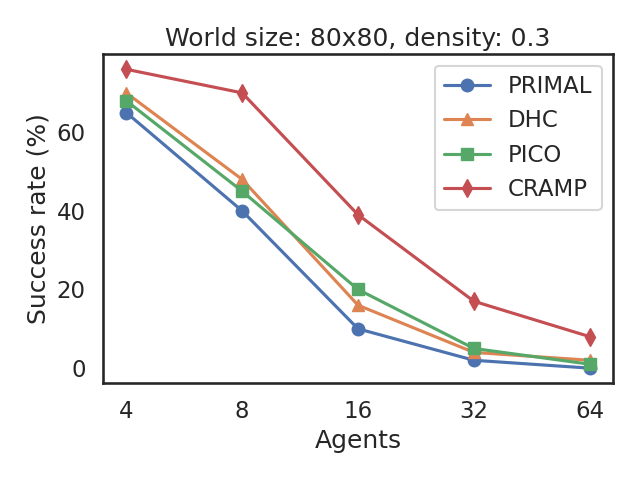}
        \label{fig:world80}
    \end{minipage}
    \vspace{-10pt}
    \caption{Performance comparison between CRAMP and other methods in larger and denser worlds.}
    \label{fig:result_comparison}
\end{figure}

To evaluate our model in larger and denser environments, we conducted experiments utilizing expanded world dimensions of $40 \times 40$ and $80 \times 80$, maintaining a consistent obstacle density of $0.3$. Comparative analysis involving CRAMP and other methods listed in Table \ref{table:results} was performed, excluding CPL due to unavailability of their code and model. The results are presented in Fig. \ref{fig:result_comparison}.

As illustrated in Fig. \ref{fig:result_comparison}, CRAMP outperforms other methods in both settings. In the $40 \times 40$ environments, CRAMP attains the success rate of 93\% for 4 agents, while other methods achieve approximately 85\%. Although success rates decrease with an increasing number of agents across all methods, CRAMP consistently maintains its lead. A parallel trend is observed within the $80 \times 80$ environments. Despite decreased success rates due to expanded search space, CRAMP maintains a substantial lead over alternative methods.

\subsection{Ablation study}

\begin{figure}[ht!]
    \centering
    \includegraphics[width=.9\columnwidth]{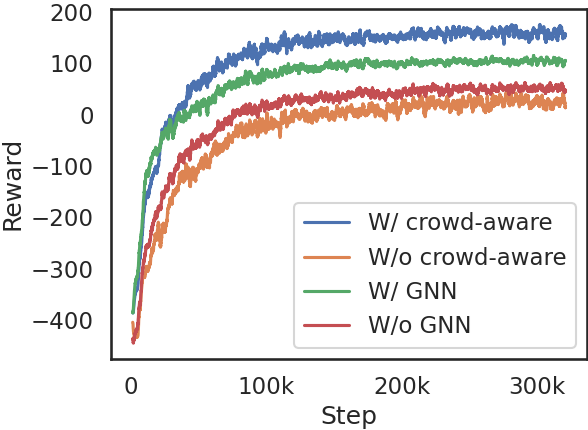}
    \caption{Effect of a crowd-aware reward function and GNN-based local communication.}
    \label{fig:ablation_reward}
\end{figure}

To examine the impact of the crowd-aware reward function and local communication via GNNs, we conduct an ablation study. This study involves individually activating and deactivating these features to observe their influence on performance, as indicated by the episode rewards collected. The results shown in Fig. \ref{fig:ablation_reward} demonstrate that the crowd-aware reward function significantly enhances model convergence. Specifically, models utilizing the crowd-aware reward achieve much quicker convergence, reaching approximately $120$ rewards per episode, whereas models lacking this feature only converge to about $-50$ rewards per episode. Similarly, incorporating GNNs positively influences convergence, with models enabled with GNNs reaching $50$ rewards per episode compared to those without GNNs, which fall below $0$ reward.

\begin{table}[ht!]
\centering
\caption{Evaluation in $80 \times 80$ environments with density of $0.1$, measured by success rate (SR) and makespan (MS)}
\label{table:ablation}
\begin{tabular}{|c|c|c|c|c|c|c|}
\hline
\multirow{2}{*}{Agents} & \multicolumn{2}{l|}{W/o crowd-aware} & \multicolumn{2}{l|}{W/o GNN} & \multicolumn{2}{l|}{CRAMP full} \\
\cline{2-7}
                            & SR           & MS           & SR        & MS       & SR         & MS         \\
\hline
4                           & 75             & 99.78              & 79         & 96.54          & \textbf{85}          & \textbf{67.34} \\
8                           & 60             & 143.58             & 65         & 114.32         & \textbf{76}          & \textbf{98.15} \\
16                          & 47             & 280.02             & 50         & 266.47         & \textbf{57}          & \textbf{225.14} \\
32                          & 9              & 543.82             & 12         & 502.12         & \textbf{18}          & \textbf{407.56} \\
64                          & 3              & 913.07             & 5          & 873.59         & \textbf{9}           & \textbf{746.83} \\
\hline
\end{tabular}
\end{table}

We conducted additional experiments within $80 \times 80$ environments with a density of $0.1$ to assess the effectiveness of our method. The findings, presented in Table \ref{table:ablation}, showcase significant enhancements in CRAMP's performance upon the integration of crowd-aware rewards and GNN-based local communication. Enabling the crowd-aware reward function consistently boosts the success rate across various agent configurations. For instance, an increase of 16\% in the success rate and a reduction of 45.43 in average makespan for 8 agents highlighting its vital role in promoting efficient navigation. Similarly, although to a slightly lesser degree, activating GNN-based local communication also yields improvements in the success rate, with a 7\% increase observed for 16 agents, implying its beneficial impact on agent coordination.

\section{CONCLUSION}

We proposed CRAMP, a novel reinforcement learning framework for Multi-Agent Path Finding (MAPF) tasks. The innovation of CRAMP lies in the carefully designed crowd-aware reward function and our GNN-based local communication mechanism. We optimize our model using a boosted curriculum training strategy. Our extensive experiments demonstrate CRAMP's capability by consistently outperforming other leading multi-agent reinforcement learning approaches across various metrics. Our method significantly enhances the quality of solutions in all tested configurations, making it a promising contribution to multi-agent path planning. Our approach can be applied to real-world applications, from autonomous robotics to intelligent transportation systems, where efficiency and safety are critical constraints.

\textbf{Limitations:} Our method may not prevent deadlocks when two robots approach the same corridor from opposite ends. Addressing such deadlocks could require alternative strategies, like prioritizing one robot's passage over the other. Additionally, due to limited computational resources, we could only train up to eight distributed agents simultaneously, resulting in worse performance when the number of agents is large. Scaling our method for a larger number of agents may benefit from the inclusion of additional agents or dummy agents in the simulation environment.









{\small
\bibliographystyle{IEEEtran}
\bibliography{refs}
}

\end{document}